\documentclass{SIW}

\usepackage[OT1]{fontenc}
\usepackage{graphicx,xcolor}
\usepackage{lipsum}
\usepackage{amsmath}
\usepackage{amsfonts}
\usepackage{amssymb,bm}
\usepackage{comment} 
\usepackage{multirow}
\usepackage{float}

\author{
	Mattia Pugliatti\textsuperscript{1}\textsuperscript{*}
	and 
	Francesco Topputo\textsuperscript{1};
	\textsuperscript{1} Politecnico di Milano, Department of Aerospace Science and Technology, Via La Masa 34, 20156, Milan, Italy.
	\textsuperscript{*}[mattia.pugliatti@polimi.it]
}

\title{Boulders Identification on Small Bodies Under Varying Illumination Conditions}

\event{Space Imaging Workshop. Atlanta, GA. \\ 10-12 October 2022}

\begin{document}

\maketitle


\MakeAbstract{The capability to detect boulders on the surface of small bodies is beneficial for vision-based applications such as navigation and hazard detection during critical operations. This task is challenging due to the wide assortment of irregular shapes, the characteristics of the boulders population, and the rapid variability in the illumination conditions. The authors address this challenge by designing a multi-step training approach to develop a data-driven image processing pipeline to robustly detect and segment boulders scattered over the surface of a small body. Due to the limited availability of labeled image-mask pairs, the developed methodology is supported by two artificial environments designed in Blender specifically for this work. These are used to generate a large amount of synthetic image-label sets, which are made publicly available to the image processing community. The methodology presented addresses the challenges of varying illumination conditions, irregular shapes, fast training time, extensive exploration of the architecture design space, and domain gap between synthetic and real images from previously flown missions. The performance of the developed image processing pipeline is tested both on synthetic and real images, exhibiting good performances, and high generalization capabilities
}

\section*{Introduction}

Missions towards small bodies, such as asteroids and comets, are becoming increasingly interesting for national space agencies, companies, and smaller players such as research centers and universities \cite{GNC_survey_JPL} . These bodies display great variability in terms of shapes and surface morphological characteristics, which pose new challenges for optical-based systems, especially under varying illumination conditions. 
Within the broader field of spacecraft autonomy, the specific capability to navigate around a known celestial body and comprehend its surroundings is of paramount importance to enable any autonomous decision-making process onboard a spacecraft. When considering the proximity environment of a small body and all the sensors available on the market, cameras are usually preferred as they are light, compact, and have low power demand. For these reasons, the use of passive cameras, in combination with Image Processing (IP) algorithms, provides compelling performance with cost-effective hardware. The inclusion of such an understanding of the surrounding environment enables autonomous systems to operate at a fraction of the cost that would be traditionally required to operate with human-in-the-loop approaches, also unlocking the capability to perform enhanced critical operations unaffected by time-delayed communications\cite{GNC_survey_JPL} . 

Within this context, robust identification of boulders on the surface of celestial bodies has implications both for features-based navigation techniques\cite{song2022deep,villaautonomous,Pugliatti2021_navseg} and for hazard detection and avoidance for landing applications\cite{Furfaro2020_DRL_Moon,DRL_Japan,Skinner2021,Caroselli2022,claudet2022benchmark}. In this work, this is approached as a semantic segmentation task.

Previous works exist in the literature that uses image segmentation in space-related applications. In \cite{Thompson2005,Wagstaff2014,goh2022mars,Palafox2017} it is used as a means to classify and distinguish geological properties of the terrain while in \cite{Fuchs2015} to enhance scientific return during flybys. With the progress of artificial intelligence, and in particular deep-learning, architectures for image-segmentation applications have boomed. Most notably, outside the space domain, in \cite{UNet} a new successful architecture referred to U-shaped Network (UNet) is introduced for biomedical segmentation. This architecture has later been extensively used for its design and implementation simplicity in works such as in \cite{Furfaro2020_DRL_Moon,DRL_Japan,Skinner2021,Caroselli2022,claudet2022benchmark} to generate hazard maps for safe landing site selections on the surface of celestial bodies such as the Moon, Mars, and small bodies. All the aforementioned works based on UNet architectures suffer the disadvantages typical of deep-learning architectures: computationally expensive networks which require a large amount of time for training and a large amount of labeled data. The latter poses a critical issue for the development and benchmark within the IP community of any data-driven algorithm \cite{song2022deep}. Previous work by the authors of this paper has also been performed about semantic segmentation on the surface of small bodies. In particular in \cite{Pugliatti2022_unet}, where a set of UNet architectures are designed to map the image content of small bodies into 5 different classes: background, surface, craters, boulders, and terminator region. These segmentation maps have also demonstrated to be usable for optical navigation in \cite{Pugliatti2021_navseg}.

An opposite trend exists in parallel that pushes for the use of shallower networks, which exploit randomization together with alternative training strategies in order to expedite the training and reduce the burden of computational complexity. Within these premises, Extreme Learning Machine (ELM) \cite{Huang2006,Huang2014,Huang2015_ELM} and its subsequent evolution to hierarchical pooling architectures referred to Convolutional Extreme Learning Machine (CELM) \cite{Huang2015_CELM,saxe,Rodrigues2021,Pugliatti2022_CELM_preprint} , have proposed a possible solution with limited degradation in performance compared to deep counterparts such as neural networks and convolutional neural networks. 

The work presented in this paper is a spin-off between the one in \cite{Pugliatti2022_unet} and the one in \cite{Pugliatti2022_CELM_preprint} with the three main following innovations. First, contrary to \cite{Pugliatti2022_unet} which focuses on five layers, in this work the focus is solely put on the boulders and their robust segmentation in face of varying illumination conditions. Second, concepts from CELM theory \cite{Huang2015_CELM,saxe,Pugliatti2022_CELM_preprint} that have been successfully tested for other IP tasks, are extended to use for image segmentation to speed up the design of the final IP pipeline. Third, particular care is set to the design of large synthetic datasets of image-label pairs through realistic artificial environments. In a push to encourage the IP community to benchmark different approaches together, these datasets are made publicly available \cite{DatasetZenodoBoulders} while their statistics are discussed in detail in \cite{DatasetPugliatti2022} . Exploiting these three pillars, the work presented in this paper outline the design of an efficient IP pipeline for robust boulder segmentation on the surface of small bodies under varying illumination conditions. 

The rest of the paper is organized as follows. First, the methodology is discussed, with particular care on the dataset's key characteristics and the design of the IP pipeline used for segmentation. Then, the results of different sections of the pipeline are presented. Finally, some conclusions and future works are discussed. 

\section*{Methodology}
The methodology is divided into two subsections, the first addressing the high-level characteristics of the datasets used in this work and the second illustrating the multi-step training strategy that produces the IP pipeline to segment boulders.

\subsection*{Datasets}

The data-driven IP pipeline designed in this work necessarily requires extensively annotated datasets about boulders on the surface of small bodies seen from varying illumination conditions. Since the remarkable lack of publicly available datasets\cite{song2022deep} from previously flown missions, artificial environments are specifically designed in this work to generate large amounts of synthetic labeled images. To do so, Blender\footnote{https://www.blender.org/, retrieved 13th of September, 2022.} is used due to its simplicity, extensive prior usage, large support community, and open-source licensing. Using these artificial environments, three main datasets have been generated specifically for this work, for simplicity referred to as $DS_1$, $DS_2$, and $DS_3$. A detailed description of the setup used to generate them as well as a statistics characterization is presented in detail in\cite{DatasetPugliatti2022} . This section focuses on a brief overview of the main characteristics of each dataset. 

$DS_1$ is composed of synthetic images of single instances of boulders positioned on a procedural-varying quasi-spherical surface. Its main purpose is to represent a single instance of a boulder positioned on the surface of a generic small body. $DS_2$ is also composed of synthetic images, however, multiple instances of boulders are scattered across the surface of an enhanced shape model of the primary body of the (65803) Didymos asteroid. This is done to represent realistic boulder distributions scattered across a generic regular shape body. Finally, $DS_3$ is composed of a small set of real images manually labeled from previously flown missions toward asteroids (25143) Itokawa\cite{Hayabusa1}, (162173) Ryugu\cite{Hayabusa2}, and (101955) Bennu\cite{OsirisRex}. Its purpose is to represent real boulder populations scattered across the surface of existing small bodies. 

\begin{figure}[h]
    \centering
    \includegraphics[width=0.49\textwidth]{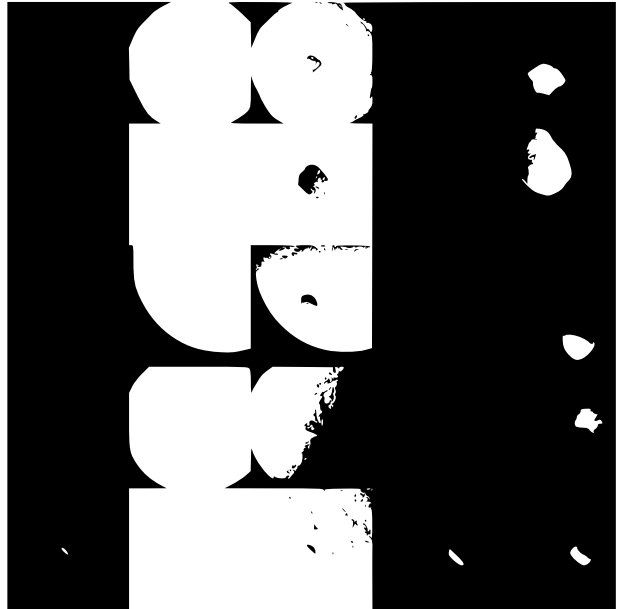}
    \caption{Sample of image-label pairs of $DS_1$. From left to right: $256\times256$ grayscale rendering in Blender, masks without shadows, masks with shadows, followed by $128\times128$ noisy and randomly cropped grayscale images, and relative masks with shadows.}
    \label{fig:DS1_sample}
\end{figure}

The image-label pairs of $DS_1$ are created using a unitary radius high-resolution spherical mesh to represent the small body, while boulder meshes are generated randomly using the \textit{Rock generator} add-on in Blender. A set of 30 meshes is used to represent archetype shapes of boulders, which are then singularly positioned on the surface of the body with random orientation, scaling, and albedo. In order to simulate camera positions, a random cloud of points is generated around the boulder, while illumination conditions are also varied randomly. During acquisition the attitude is assumed to be ideal, pointing towards the center of the boulder. Images are rendered at a resolution of $256\times256$, but are then post-processed with random cropping to $128\times128$ size images and the addition of artificial noise. Post-processing is fundamental in making sure the boulders are not always centered in the images. Both boulders and surfaces are simulated utilizing an Akimov scattering law implemented in the shading tab of Blender. At each acquisition, the characteristics of the surface of the body are varied randomly to simulate different roughness that disturbs the environment surrounding each boulder. With this setup, a total of $45269$ image-label pairs are rendered for $DS_1$. Note that the masks of the boulder and surface are obtained thanks to the \textit{Cycles} rendering engine in Blender and are generated both with and without shadows. These are later split into training, validation, and test sets as illustrated in Table \ref{tab:Datasets_summary}. Figure \ref{fig:DS1_sample} represents a sample of image-label pairs of $DS_1$ after rendering and after post-processing.

\begin{figure}[h]
    \centering
    \includegraphics[width=0.25\textwidth]{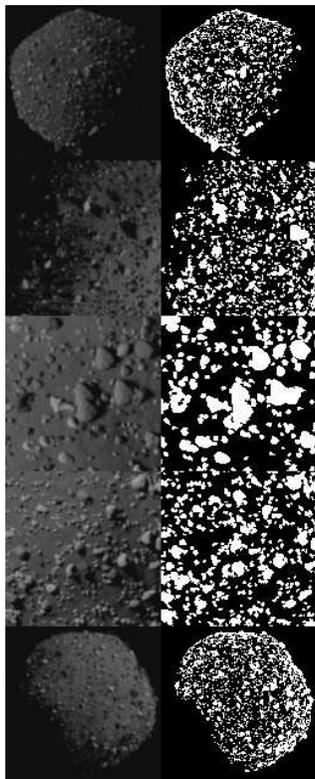}
    \caption{Sample of image-label pairs of $DS_2$. $128\times128$ noisy grayscale images (left) and relative boulder masks (right).}
    \label{fig:DS2_sample}
\end{figure}

The procedure adopted to generate the image-label pairs of $DS_2$ is in part similar to the one illustrated for $DS_1$. The main differences are in the number of boulders positioned on the surface of the body, the size of the rendered images in Blender ($128\times128$), and the lack of random cropping during post-processing (only artificial noise is added to the rendered images). During rendering, instead of placing a single boulder, multiple ones are positioned on the surface of the enhanced Didymos shape model to represent a realistic boulders distribution. Once again, as in $DS_1$, both lighting, scale, albedo, and intensity variations are randomly implemented to obtain a generalized dataset. $DS_2$ is made of $35183$ image-label pairs. Their split into training, validation, and test sets is summarized in Table \ref{tab:Datasets_summary}, while a sample of image-label pairs is visible in Figure \ref{fig:DS2_sample}.

\begin{figure}[h]
    \centering
    \includegraphics[width=0.25\textwidth]{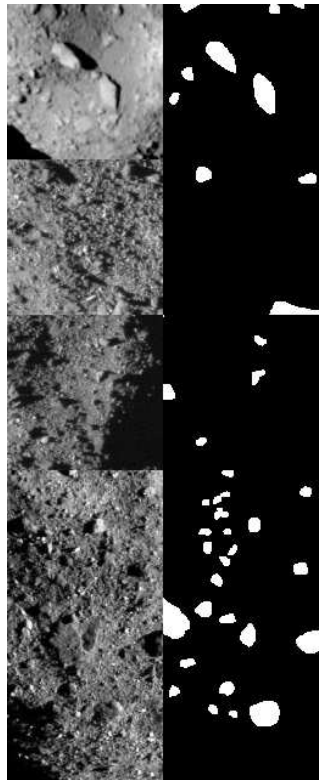}
    \caption{Sample of image-label pairs of $DS_3$. $128\times128$ real grayscale images (left) and relative boulder masks (right), manually labeled in \cite{Pugliatti2022_unet}.}
    \label{fig:DS3_sample}
\end{figure}

Finally, the $DS_3$ dataset is generated starting from $75$, $256\times256$ cropped images which show clear boulders presence that has been manually labeled in \cite{Pugliatti2022_unet} . Each image-mask pair is then subdivided into 4 $128\times128$ smaller ones to reach a total of $300$ samples. By design, this dataset only contains the masks of the largest boulders, as is visible in the sample in Figure \ref{fig:DS3_sample}.

Table \ref{tab:Datasets_summary} summarizes the main characteristics of the split used in the train, validation, and test sets used in this work. Note that, as explained in \cite{DatasetPugliatti2022} , the main difference between $Te_1$ and $Te_2$ is that the first represents images with a balanced distribution of phase angles while the latter does not.

\begin{table}[]
    \centering
    \begin{tabular}{rrrrr}
        \hline
        \hline
        Name & Acronym & $DS_1$ & $DS_2$ & $DS_3$ \\
        \hline
        Training & $Tr$ & 30181 & 20095 & - \\
        Validation & $V$ & 5044 & 5044 & -\\
        \multirow{3}{*}{Test} 
         & $Te_1$ & 5044 & 5044 & - \\
         & $Te_2$ & 5000 & 5000 & - \\
         & $Te_3$ & - & - & 300 \\
        \hline
        Total & - & 45269 & 35183 & 300
    \end{tabular}
    \caption{Summary of the datasets used in this work.}
    \label{tab:Datasets_summary}
\end{table}

\subsection*{Network architectures}

\begin{figure}[h!]
	\centering
	\includegraphics[width=0.48\textwidth]{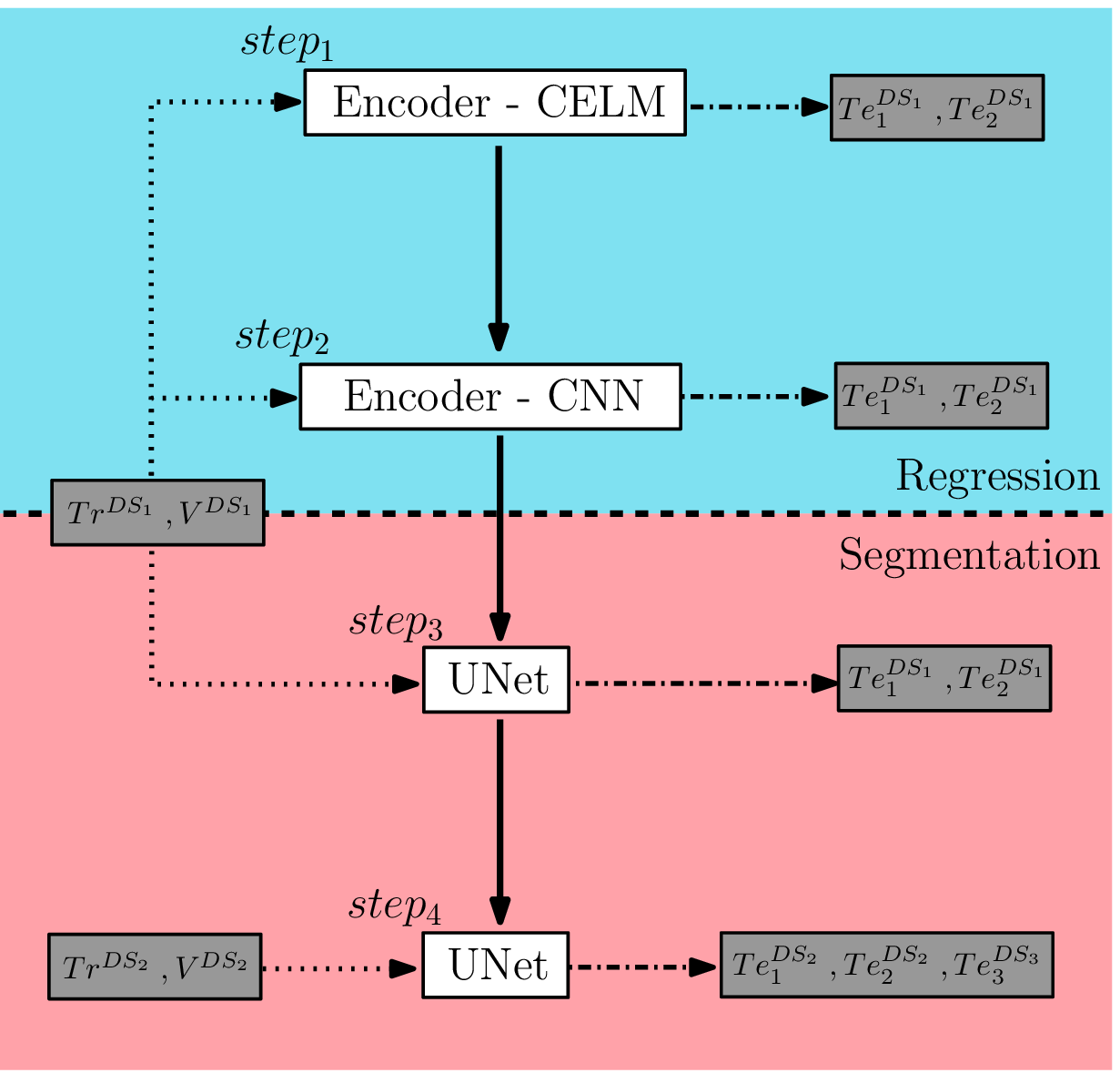}
	\caption{Schematic of the whole training procedure used in this work.}
	\label{fig:training_all}
\end{figure}

In this work, a data-driven IP pipeline is designed to perform robust boulder segmentation under varying illumination conditions. The pipeline is designed specifically for this work following an incremental training strategy that involves different architectures. The strategy is intended to efficiently accompany the design by incrementally training portions of the final architecture that performs boulders segmentation on the surface of small bodies. The training is designed as a 4 steps process, as schematized in Figure \ref{fig:training_all}, using a Tesla P100-PCIE 16Gb GPU, with a 27.3 Gb of RAM in Google colab\footnote{https://colab.research.google.com/, retrieved 13th of September, 2022.}.

In the first two steps, images from $DS_1$ are used together with the $CoB$ of the boulders masks with shadows to design an encoder structured as a Hierarchical Pooling Network (HPN). First, to select the proper architecture of the HPN, the CELM framework \cite{Huang2015_CELM,Rodrigues2021,saxe} is selected as an effective tool to efficiently explore the architecture design space. In CELM theory training of a HPN happens by solving a regularized least-square problem that finds the best set of weights $\bm{\beta}$ connecting the last hidden layer of the architecture with its output layer. All remaining weights $\bm{W}$ and biases $\bm{b}$ of the network are set randomly at initialization and are kept frozen during training. To find out $\bm{\beta}$ means to solve the following problem\cite{Huang2015_CELM} : 
\begin{equation}\label{eq:rls_prob_statement}
    \text{Minimize}:       \left\| \bm{\beta} \right\|^{2}_{2} + C\left\| \bm{H\beta}-\bm{T} \right\|^{2}_{2}
\end{equation}
where $\bm{H}_{N\times L}$ is the hidden layer output matrix generated by an input tensor of depth $N$ that passes through the network up to the last dense layer before the output layer (which in this work is the fully connected layer made by $L$ neurons), $\bm{T}_{N\times M}$ is the target output layer, where $M$ is the number of labels to predict for each image. The addition of the regularization term $C$ and the minimization of the norm of $\bm{\beta}$, is proven to increase stability and generalization of the network \cite{Huang2014}. The solution of Equation \ref{eq:rls_prob_statement} is:

\begin{equation}\label{eq:celm_ls}
    \bm{\beta} = \left\{\begin{matrix}
 \bm{H}^T\left(\frac{\bm{I}}{C} + \bm{HH}^T\right)^{-1}\bm{T}, \quad \text{if} \; N\leq L \\
 \left(\frac{\bm{I}}{C} + \bm{H}^T\bm{H}\right)^{-1}\bm{H}^T\bm{T}, \quad \text{if} \; N > L 
\end{matrix}\right.
\end{equation}
depending on whether it is more convenient to invert an $N\times N$ or $L\times L$ matrix. In this work, the validation set is used during training to select the best regularization parameter $C$, while the training set is used to determine the best set of $\bm{\beta}$. Since training happens by solving a non-iterative regularized least square problem and since a single passage of the input tensor is required to compute $\bm{H}$, training time is extremely short. Such property enables a fast and efficient exploration of the architecture design space, as illustrated in \cite{saxe,Pugliatti2022_CELM_preprint} .

\begin{figure}[h!]
	\centering
	\includegraphics[width=0.49\textwidth]{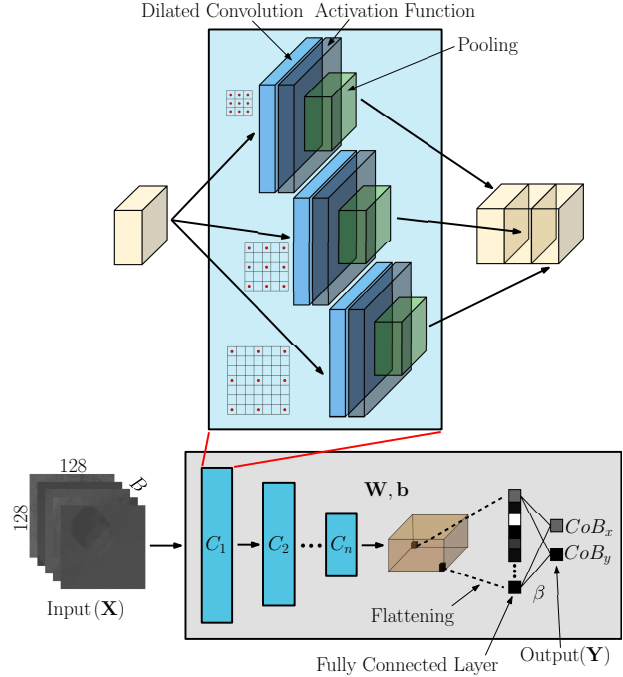}
	\caption{Schematic of the encoder used in this work.}
	\label{fig:encoder_architecture}
\end{figure}

In this work, the CELM training capabilities are exploited to select the best-performing architecture for an encoder that will be part of the final architecture. By pre-defining, a set of rules to build up a HPN as an encoder for the prediction of the boulders CoB, the architecture with the best capacity is found. As schematized in Figure \ref{fig:encoder_architecture} the encoder is designed by a sequence of cells $C_i$ which operate on batches of tensors to generate other tensors. Each cell is composed of a combination of dilated convolutions, activation functions, and pooling layers, as exemplified in the top part of Figure \ref{fig:encoder_architecture}. In particular, dilated convolutions are inserted into the encoder for their beneficial effects\cite{goh2022mars,chen2017rethinking} in augmenting the receptive field of the kernels as well as their capability to boost segmentation performance. In this work, dilated convolution with rates 1, 2, and 3 are used and then stacked together to produce the output tensor.

By assuming a constant kernel size of $3\times3$ and an exponential depth expansion coefficient equal to $2$, different architectures are generated by varying the pooling strategy $P$, the initial depth of the network $d_0$, the total number of cells $n$, the activation functions $A$, the kernel initialization strategy $K_d$, and the number of random runs for each architecture $N_r$. The regularization parameter $C$ is varied for each architecture as $10^{-3}, 10^{-2}, 10^{-1}, 1, 10^{1}, 10^{2}, 10^{3}$. By combining together these parameters a total of $1134$ different architectures are generated. The setup that achieves the best performance in predicting the boulder's CoB is represented in bold in Table \ref{tab:CELM_HPs}, which is achieved using $C = 1$.

\begin{table}[h!]
    \centering
    \begin{tabular}{p{0.7cm} p{6cm}}
        \hline
        \hline
        \textbf{Name} & \textbf{Values}\\
        \hline
        $P$ & mean, \textbf{max} \\ 
        $d_0$ & $4$, $5$, $\bm{16}$\\
        $n$ & (3,4,5)$_{d_0=4}$, (4,$\bm{5}$,6)$_{d_0=8}$, (5,6,7)$_{d_0=16}$ \\
        $A$ & NReLU, ReLU, LReLU, \textbf{ELU}, tanh, sigmoid, none \\
        $K_d$ & RandomUniform (-1,1), RandomNormal (0,1), \textbf{Orthogonal} \\
        $N_r$ & 3 \\
        \hline
    \end{tabular}
    \caption{Summary of the hyper-parameters used in this work to search for the optimal architecture of the encoder.}
    \label{tab:CELM_HPs}
\end{table}

\begin{figure*}[h!]
    \centering
    \includegraphics[width=0.8\textwidth]{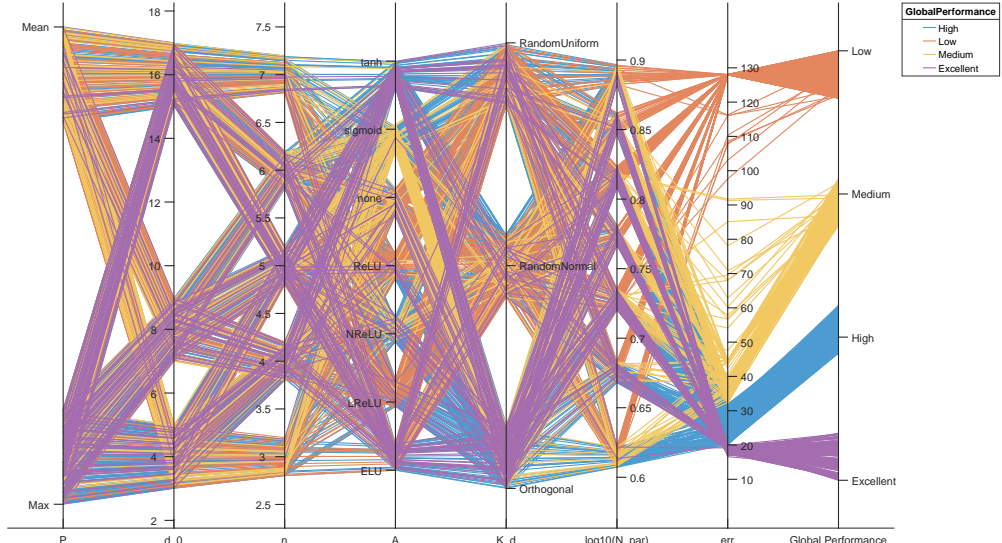}
    \caption{Parallel-plot showing the dependencies between the different hyper-parameters illustrated in Table \ref{tab:CELM_HPs} and the network performances. The lines are colored by 4 different quality metrics related to the performance (excellent, high, medium, and low).}
    \label{fig:ParallelPlot}
\end{figure*}

In Figure \ref{fig:ParallelPlot} it is possible to visualize how the combination between the hyper-parameters in Table \ref{tab:CELM_HPs} influences the performance of the HPN in its $1134$ combinations. 

In this work, the best-performing architecture out of the pool of the one tested has been chosen for implementation and is generally referred to as the CELM-encoder. However, the approach presented in this paper, coupled with representative global metrics of interest, can be used to select families of architectures that are expected to work well for a given task. The total time needed to train the $1134$ architectures with the CELM paradigm is equivalent to 48.3 hours. On average, $13.93\%$ of the time is spent on the forward pass of the validation tensor, $81.32\%$ on the forward pass of the training tensor, while the remaining $4.75\%$ is spent solving equation \ref{eq:celm_ls}. The time saved exploring the architecture design space with CELMs is crucial to fast forward and ease the training in the following steps. 

In the second step of the training, the single best performing HPN trained with the CELM paradigm is re-trained using mini-batch gradient descent\cite{Szeliski2022} such as in the case of a traditional CNN. During this training, the architectural elements are frozen while the weights and biases of the kernels are optimized starting from various batch sizes (64, 128, 256, 512) and learning rates($10^{-4}$, $10^{-3}$, $10^{-2}$). Each setup is initialized and run randomly 2 times for a total of $24$ training instances for short epochs. The best-performing ones are selected based on the best error achieved over the validation split during the entire training and are re-run while increasing the epochs. The loss function used to train the network is the mean squared error. The final setup is achieved using a batch size $B$ of $32$ samples, with a learning rate $lr$ of $10^{-4}$, and a dropout rate in the fully connected layer of $0.2$. 

\begin{figure*}[h!]
	\centering
	\includegraphics[width=0.9\textwidth]{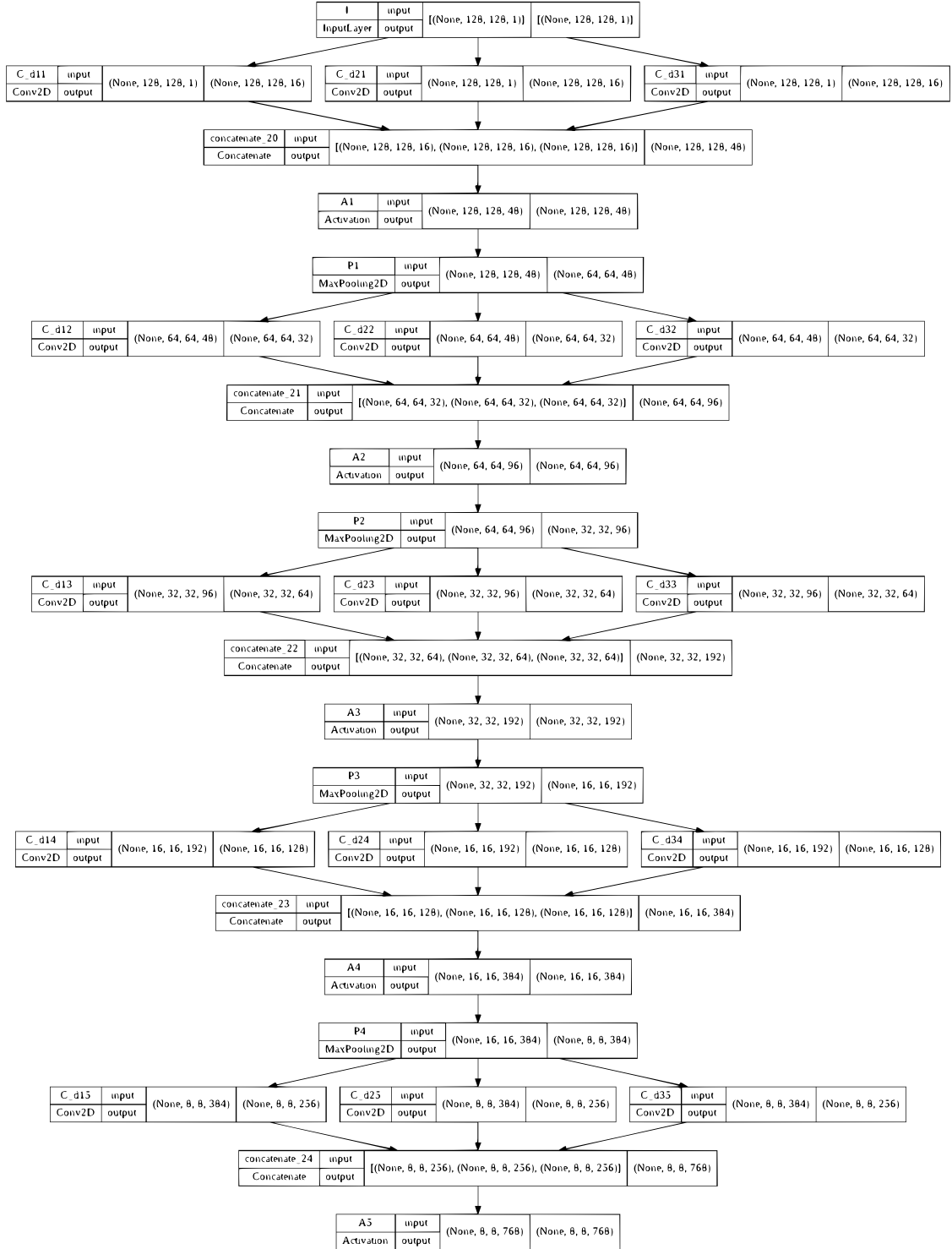}
	\caption{Detailed architecture of the encoder, made of 3'527'040 parameters.}
	\label{fig:encoder_architecture_functional}
\end{figure*}

Note that the training time of the best performing architecture for $200$ epochs using the mini-batch gradient descent method required a total of $9504.7 s$, while the equivalent training time with a CELM would take on average 153s per architecture (spent mostly in the forward pass to generate $\bm{H}$ for the training and validation sets). The total training time to find out the best combination of learning rate and batch size in \textit{step$_2$} is roughly 24h. Should all the $1134$ architectures have been trained using the mini-batch gradient descent, considering the encoder as a CNN, the exploration of the architecture design space would have resulted in a much more computationally expensive process. The combination between CELM and CNN training paradigms allows thus for efficient exploration of the architecture design space and thus of the encoder's design. The detailed architecture of the CELM and CNN encoders is represented in Figure \ref{fig:encoder_architecture_functional} using TensorFlow 2.10 notation. 

\begin{figure}[h!]
	\centering
	\includegraphics[width=0.49\textwidth]{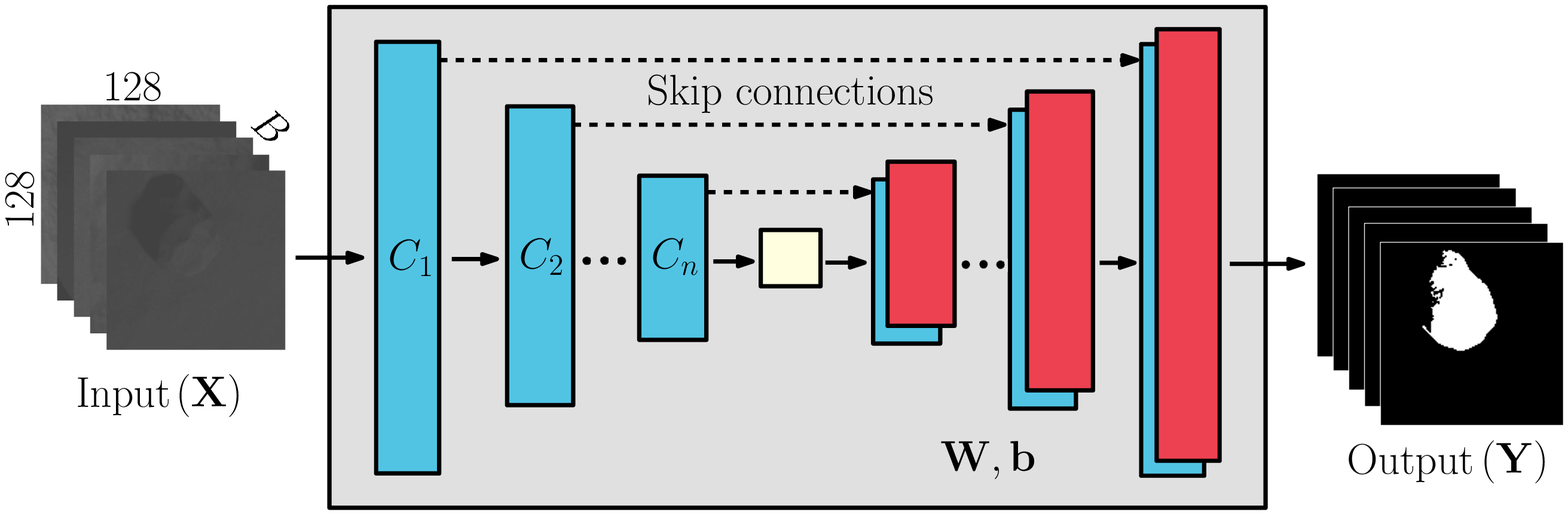}
	\caption{Schematic of the UNet architecture used in this work.}
	\label{fig:unet_architecture}
\end{figure}

The third step of the training exploits the CNN-encoder refined from the previous step and inserts it into a larger architecture configured for segmentation. For such a task, a UNet \cite{UNet} is considered, as it has shown good performance and generalization capabilities\cite{Pugliatti2022_unet} with small body images. A schematic of the UNet architecture is represented in Figure \ref{fig:unet_architecture}. Its main characteristics are: an encoder-decoder setup that resembles a U-shape, the lack of a fully connected layer, the presence of skip connections after the activation layer of each cell which are copied and concatenated to their corresponding layers in the decoder, and the fact that training only involves the decoder while the encoder is kept frozen as it already possess encoding capabilities acquired from previous training on other tasks.

\begin{figure*}[h!]
	\centering
	\includegraphics[width=0.8\textwidth]{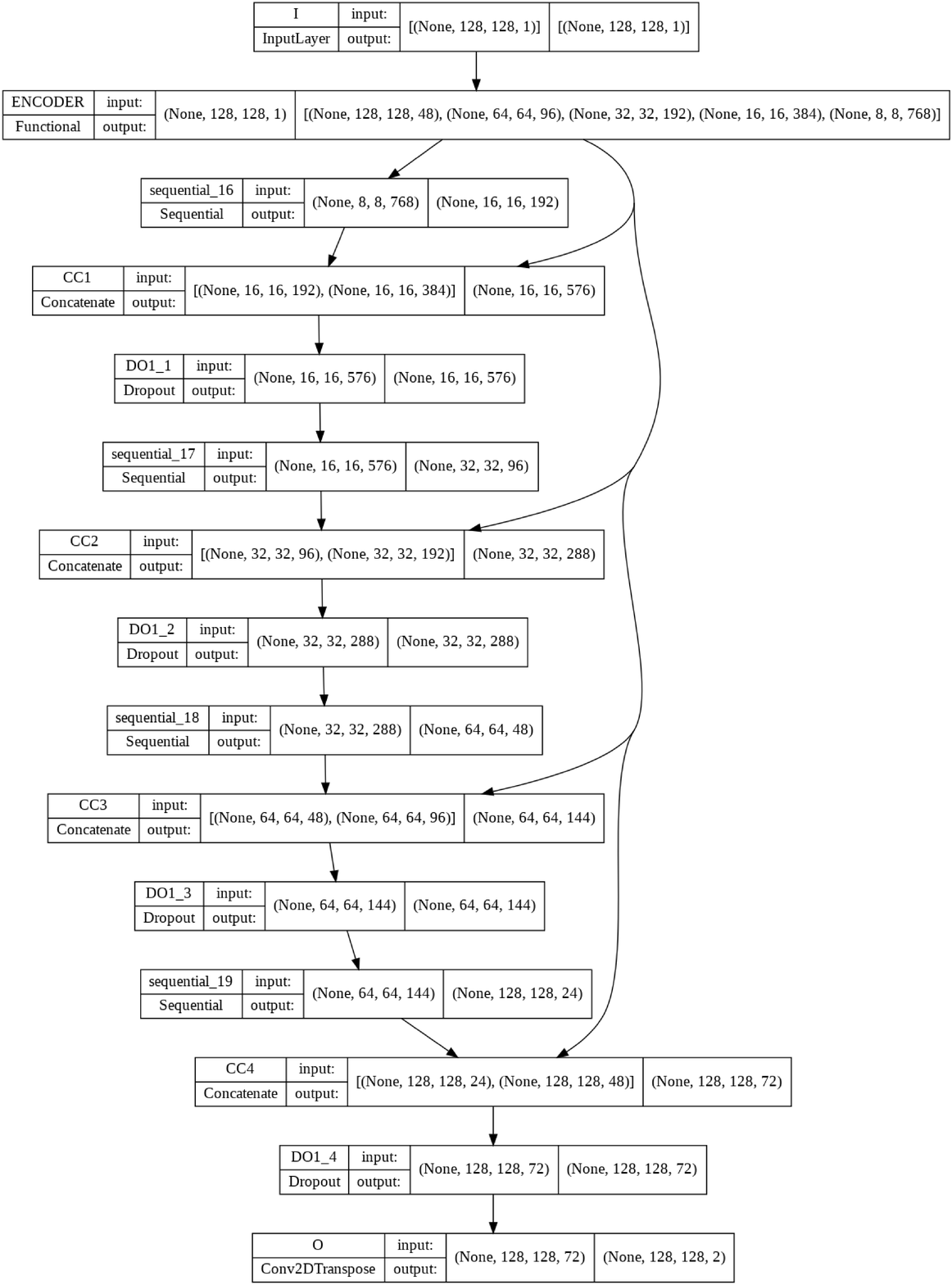}
	\caption{Detailed architecture of the UNet, made of 5'510'066 parameters, 1'982'306 of which are trainable.}
	\label{fig:unet_architecture_functional}
\end{figure*}

Similarly to \cite{Pugliatti2022_unet}, the UNet is trained by incrementally increasing the epochs while testing various combinations of dropout values, batch size, learning rate, and depth of the decoder layers. A Weighted Sparse Categorical Cross Entropy (WSCCE) \cite{Pugliatti2022_unet} is used as a loss function, while the Mean Intersection Over Union (MIOU) is used as a metric. The weights for the WSCCE loss are computed from statistical analysis of the pixel content in the masks of the training set of $DS_1$. As $3.99\%$ of the pixels are boulders while $93.01\%$ are not, the complement of these values are used respectively as weights of the non-boulder and boulder classes. The best performing architecture has been found with a dropout equal to $0.2$, a batch of $256$, a learning rate of $0.001$, and decoder depths of $192, 96, 48, \text{and } 24$. The total training time spent to find out this setup is equal to 18.9 hours. The detailed UNet architecture is illustrated in Figure \ref{fig:unet_architecture_functional} using TensorFlow 2.10 notation. 

Finally, in the fourth and last step of the training the same procedure illustrated in the previous step is repeated but using the $DS_2$ dataset. In this case, the training of the UNet is not initialized from scratch, but starts from the set of weights and biases found in the previous step, to help the network to plateau at higher values of the MIOU on the validation set, thus achieving better performance and generalization capabilities that it would from starting from scratch. The WSCCE loss function uses weights corresponding to $27.52\%$ and $72.48\%$ respectively for the non-boulder and boulder classes. The final architecture is trained over 400 epochs with a learning rate of $0.001$ and a batch size of $16$. The total training time spent to find out this setup is equal to 12 hours.

\begin{figure}[h!]
    \centering
    \includegraphics[width=0.49\textwidth]{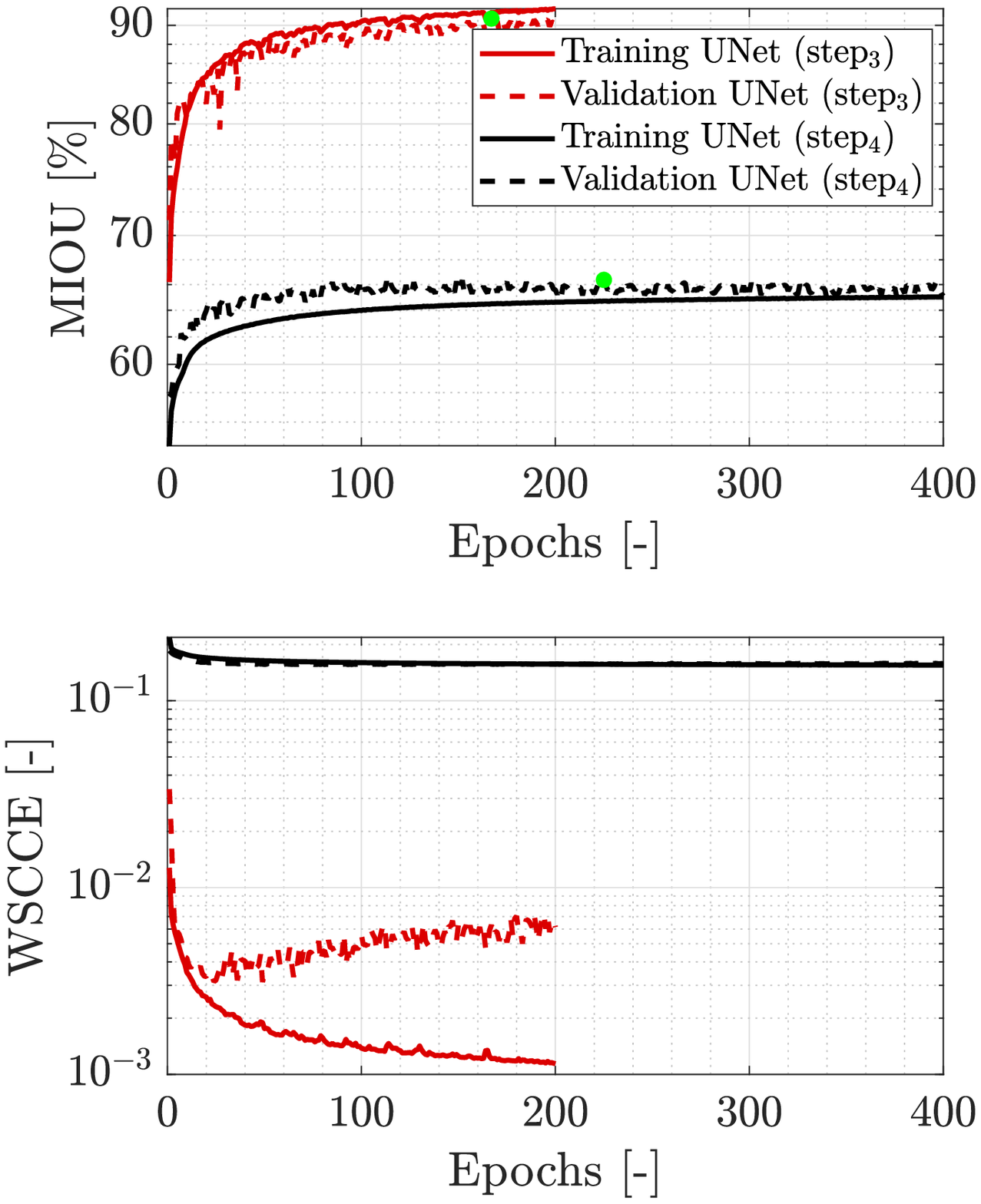}
    \caption{Training history of the MIOU (top) and WSCCE (bottom) of the UNet trained in \textit{step$_3$} and \textit{step$_4$}.}
    \label{fig:training_history_UNET}
\end{figure}

Figure \ref{fig:training_history_UNET} illustrates the entire training history of the final UNets trained in steps 3 and 4. The green points represent the events in which the maximum value of MIOU has been achieved over the validation set. When the architectures are initialized for testing, they are loaded with the sets of weights and biases that have been achieved at these epochs. It is noted that the training of the UNet with single boulders required roughly 5 hours while the one with multiple boulders 3 hours.

A schematic that summarizes the entire 4-step training procedure is illustrated in Figure \ref{fig:training_all}. CELM theory is used in \textit{step$_1$} to expedite the architecture design search of an encoder, which is further refined in \textit{step$_2$} as a CNN. The encoder is trained over a regression task on the $DS_1$ dataset, to predict the CoB of a single boulder appearing in the image. A partial training of the UNet for segmentation is then performed in \textit{step$_3$} using $DS_1$, which is further refined in \textit{step$_4$} with the use of the $DS_2$ dataset. This incremental approach ultimately allows better generalization and improved performance of the final UNet when compared to a training executed from scratch considering only \textit{step$_4$}\cite{Pugliatti2022_unet} . The entire training, from top to bottom, to obtain the final IP network for segmentation took roughly 103 hours. 

\section*{Results}
In this section, the results of the IP networks are illustrated and discussed in detail. First, the performances of the encoders are commented, followed up by those of the UNet architectures on the single and multiple boulder cases. 

\subsection*{Encoder networks}
To train the encoder and to define its generalization capabilities on the validation and test sets, an error metric is defined as: 

\begin{equation}
    \varepsilon_{CoB} = \sqrt{\left(CoB^{u}_{e}-CoB^{u}_{t}\right)^2+\left(CoB^{v}_{e}-CoB^{v}_{t}\right)^2}
\end{equation}
where $CoB^{u}_{e}$ and $CoB^{v}_{e}$ are the estimated coordinates of the CoB respectively in the $u$ and $v$ axes of the image plane and $CoB^{u}_{t}$ and $CoB^{v}_{t}$ are the corresponding true coordinates.

The performances of the CELM and CNN encoders designed in this work are illustrated in Table \ref{tab:regression_performance} for the two test sets of $DS_1$.

\begin{table}[h]
    \centering
    \begin{tabular}{c c c c}
        \hline
        \hline
        Encoder & Dataset &  $\mu(\varepsilon_{CoB})\;[px]$ & $\sigma(\varepsilon_{CoB})  \;[px]$\\
        \hline
        CELM & Te1 & 16.36 & 11.46 \\ 
        CELM & Te2 & 16.30 & 11.36 \\ 
        CNN & Te1 & 7.01 & 7.30 \\ 
        CNN & Te2 & 7.06 & 7.40 \\ 
        \hline
    \end{tabular}
    \caption{Encoders performances on the test sets of $DS_1$.}
    \label{tab:regression_performance}
\end{table}

 It is possible to see that the error is in the same order of magnitude between the two encoders, the CNN being the best performing one. It is clear that the mini-batch gradient descent optimization involving all weights and biases of the kernels in the convolutional layers favors a better-performing network. Moreover, it is also observed that both encoders exhibit limited variability between $Te_1$ and $Te_2$. 

\begin{figure}[h!]
	\centering
	\includegraphics[width=0.43\textwidth]{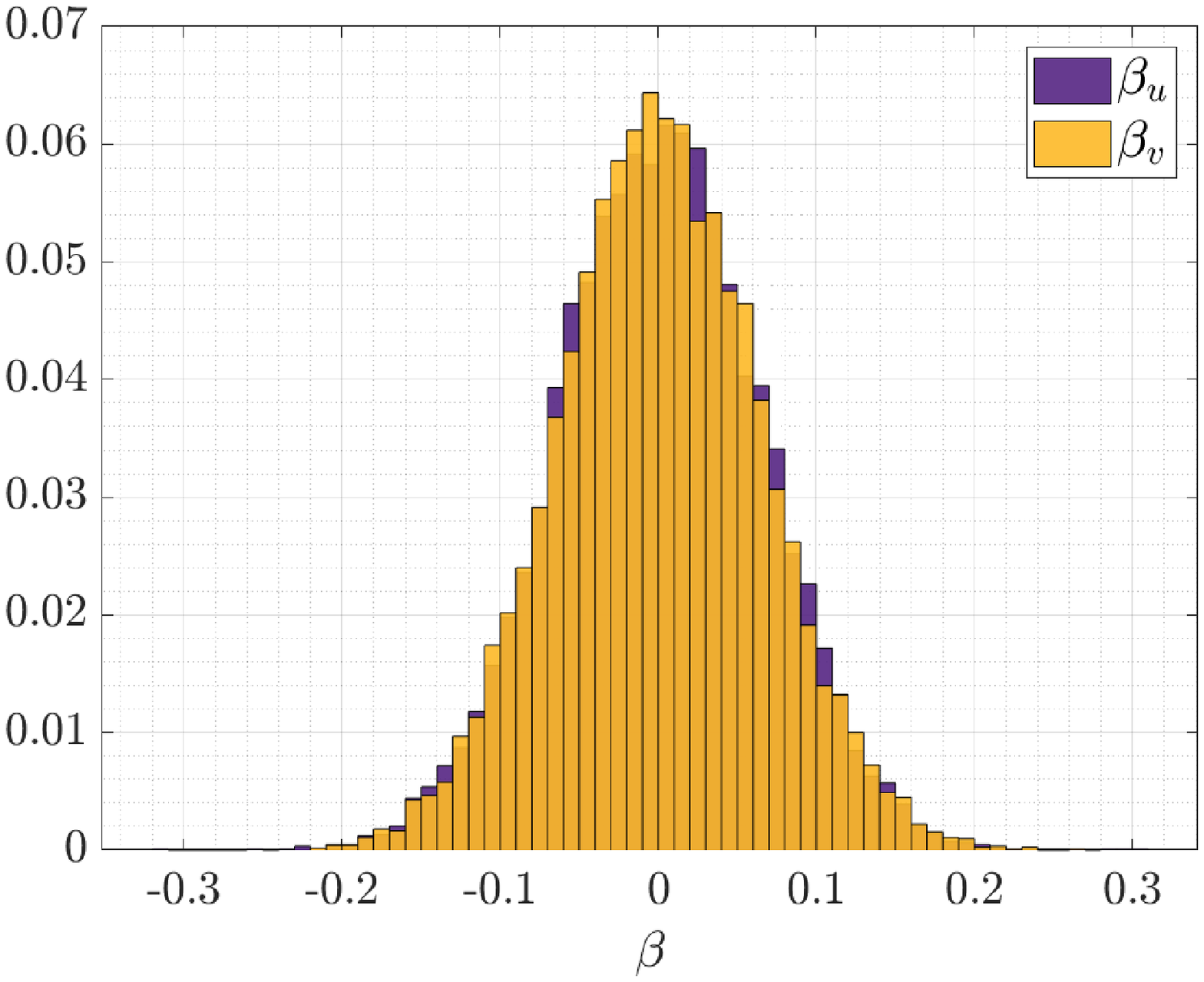}
	\caption{Distribution of the weights $\bm{\beta}$ in the last layer of the CELM-encoder. The y-axis shows the relative probability of each bin.}
	\label{fig:histogram_CELM_beta}
\end{figure}

\begin{figure}[h!]
	\centering
	\includegraphics[width=0.43\textwidth]{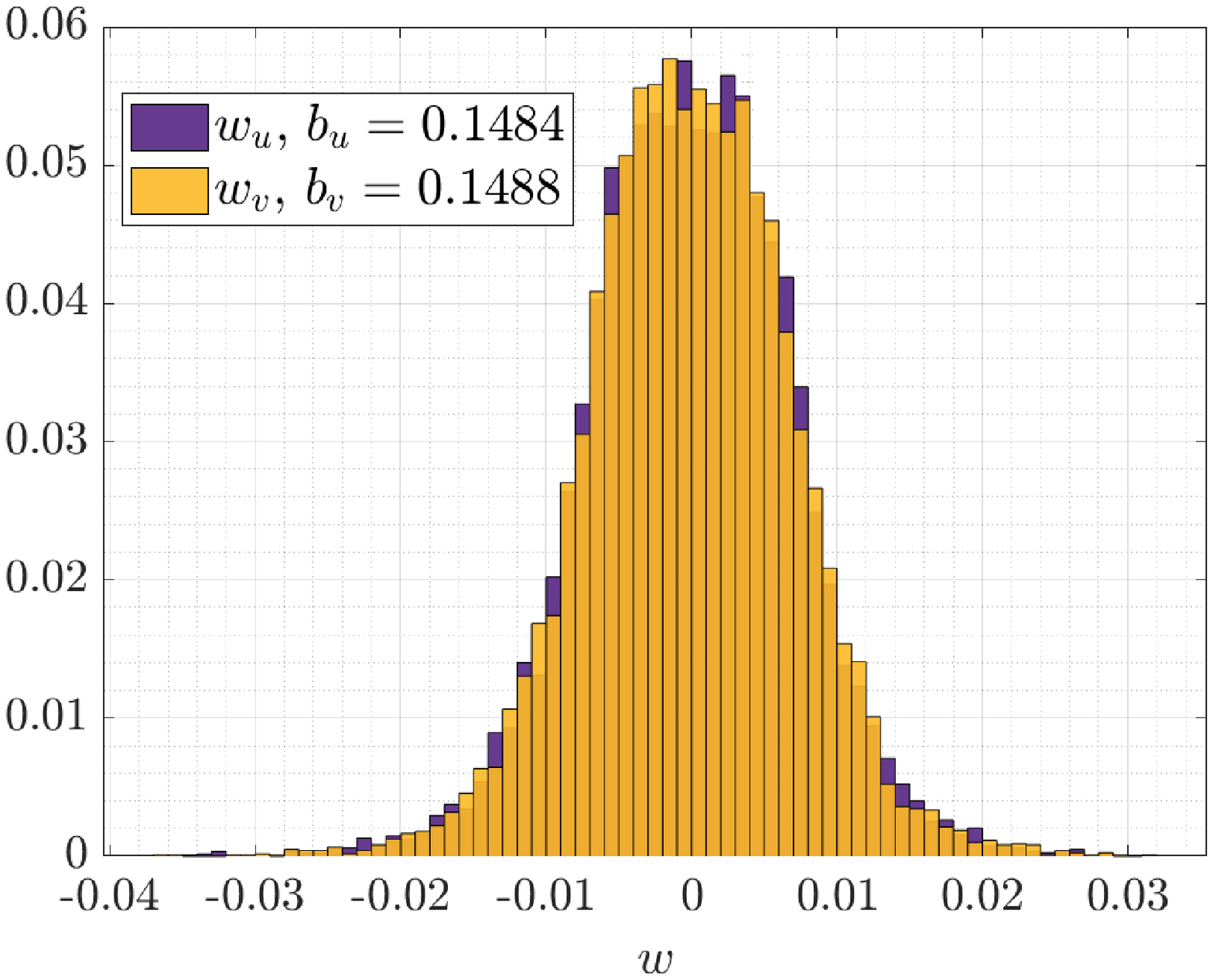}
	\caption{Distribution of the weights $\bm{w}$ in the last layer of the CNN-encoder. The y-axis shows the relative probability of each bin.}
	\label{fig:histogram_CNN_beta}
\end{figure}

In Figure \ref{fig:histogram_CELM_beta} and Figure \ref{fig:histogram_CNN_beta}, the histograms of the distributions of $\bm{\beta}_u$, $\bm{\beta}_v$, $\bm{W}_u$ and $\bm{W}_v$ between the last hidden layer and the output layer of the CELM and CNN encoders are illustrated. Both sets of weights are normally distributed, however it is noted that those of the CNN exhibit a variance that is one order of magnitude smaller. Finally, it is also observed that while the CELM encoder does not possess a bias term in the connection between the two layers, the CNN does and its value is very similar between its two components ($0.148$).

\subsection*{Segmentation networks}

The performances of the best UNet architecture generated after \textit{step$_3$} of the overall training procedure are summarized in Table \ref{tab:segmentation_UNET1_performances} on the test sets of $DS_1$ in terms of WSCCE, Accuracy (A), and MIOU. As for the encoder case, the network performs similarly in $Te_1$ and $Te_2$, exhibiting very high values of MIOU. It is noted that the values achieved by such a network are an improvement of the ones in \cite{Pugliatti2022_unet}, where the segmentation network is tasked to predict a 5-layer mask instead of the 2-layer one of this work. It is also observed that the network is capable to predict a single boulder's presence robustly under a variety of illumination conditions. 

\begin{table}[h]
    \centering
    \begin{tabular}{c ccc}
        \hline
        \hline
        Dataset &  $\mu(WSCCE)\;[-]$ & $\mu(A)\;[\%]$ & $\mu(MIOU)\;[\%]$ \\
        \hline
        Te1 & 7.4 $10^{-3}$ & 99.19 & 90.78 \\ 
        Te2 & 9.6 $10^{-3}$ & 99.20 & 91.03 \\ 
        \hline
    \end{tabular}
    \caption{UNet performance on the test sets of $DS_1$.}
    \label{tab:segmentation_UNET1_performances}
\end{table}

Similarly, Table \ref{tab:segmentation_UNET2_performances} summarizes the performances of the UNet trained in \textit{step$_4$} with images of multiple boulders on the test sets of $DS_2$ and $DS_3$. Indeed, the presence of a large population of multiple boulders seems to challenge the performance of the UNet, which is lower than the one trained with single boulders in \textit{step$_3$}. This is also illustrated by the WSCCE (which is two orders of magnitude higher than in the previous case) as well as from the values of $A$ and $MIOU$.

\begin{table}[h]
    \centering
    \begin{tabular}{c ccc}
        \hline
        \hline
        Dataset &  $\mu(WSCCE)\;[-]$ & $\mu(A)\;[\%]$ & $\mu(MIOU)\;[\%]$ \\
        \hline
        Te1 &  1.59 $10^{-1}$ & 82.22 & 66.09\\ 
        Te2 &  1.6 $10^{-1}$ & 81.98 & 66.04\\
        Te3 &  2.8 $10^{-1}$ & 62.53 & 33.26\\ 
        \hline
    \end{tabular}
    \caption{UNet performance on the test sets of $DS_2$ and $DS_3$.}
    \label{tab:segmentation_UNET2_performances}
\end{table}

It is also commented that the current masks in $DS_3$ demonstrated to be inappropriate for the specific design in this work. These masks were originally designed in \cite{Pugliatti2022_unet} by manual labeling of the most prominent boulders in $256\times256$ images. However, the UNet developed in this work is capable of predicting much more boulders than the ones represented by the ground truth masks in $DS_3$, making them unreliable for quantitative analyses. Nonetheless, images from this dataset turned out to be useful for a qualitative assessment. 

\begin{figure}[h!]
    \centering
    \includegraphics[width=0.3\textwidth]{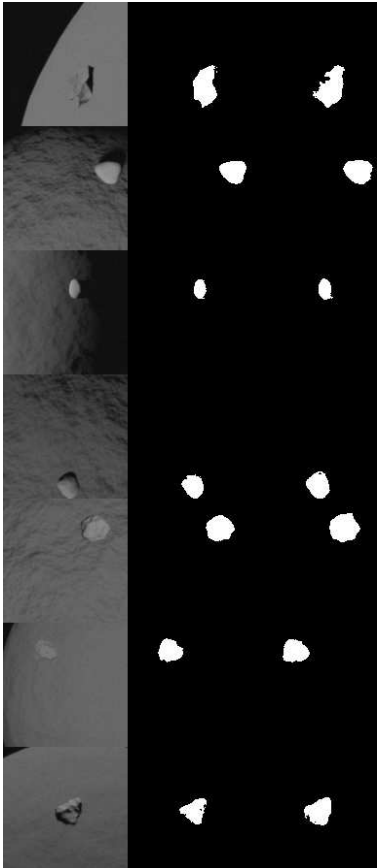}
    \caption{Samples of input image (left), true mask (center), and predicted mask (right) from the UNet trained in \textit{step$_3$} on test images of $DS_1$.}
    \label{fig:UNET1_Te1_DS1}
\end{figure}

\begin{figure}[h!]
    \centering
    \includegraphics[width=0.3\textwidth]{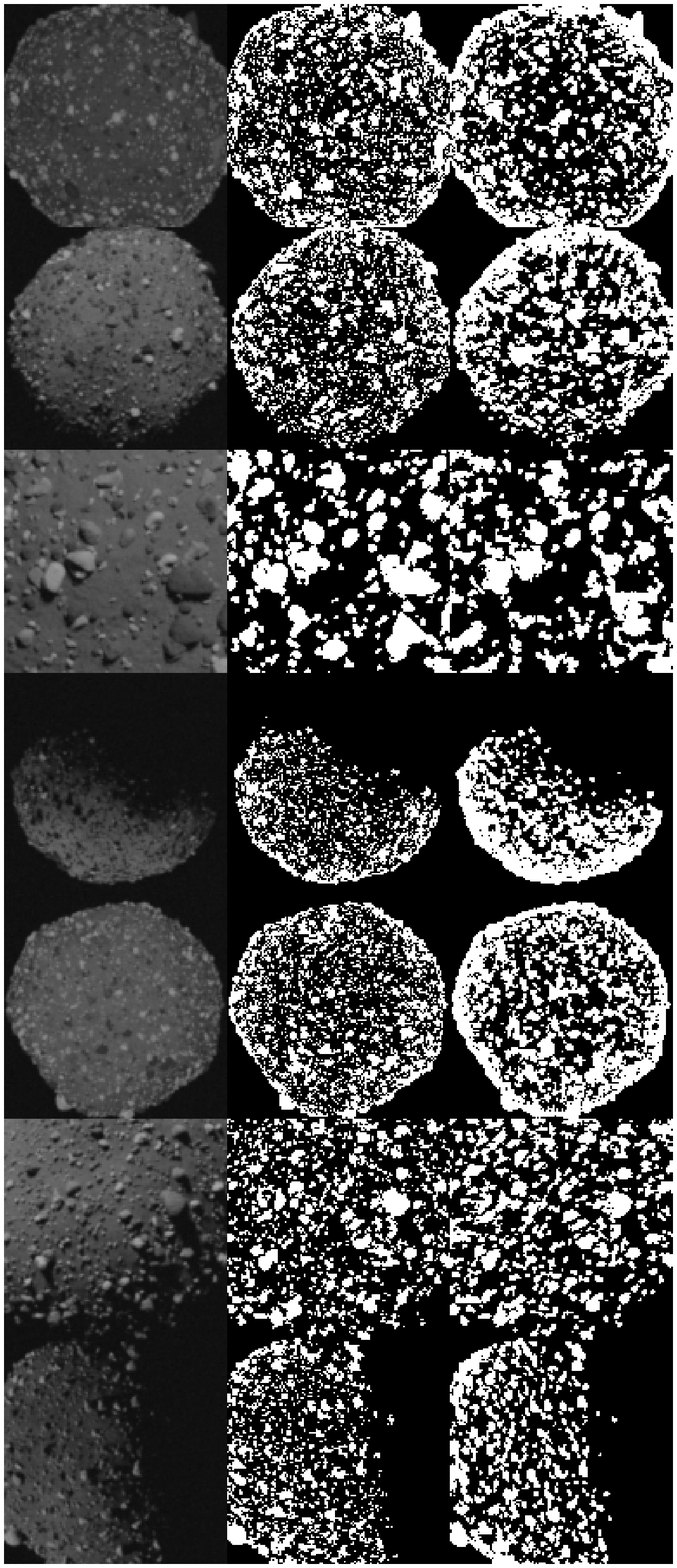}
    \caption{Samples of input image (left), true mask (center), and predicted mask (right) from the UNet trained in \textit{step$_4$} on test images of $DS_2$.}
    \label{fig:UNET2_Te1_DS2}
\end{figure}

\begin{figure}[h!]
    \centering
    \includegraphics[width=0.3\textwidth]{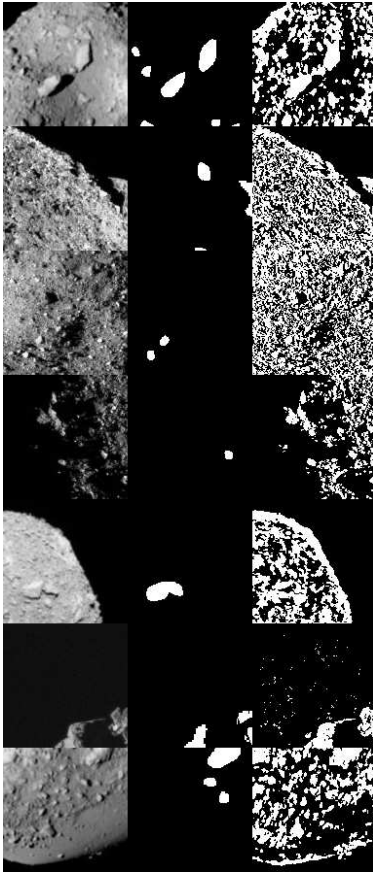}
    \caption{Samples of input image (left), true mask (center), and predicted mask (right) from the UNet trained in \textit{step$_4$} on test images of $DS_3$.}
    \label{fig:UNET2_Te1_DS3}
\end{figure}

Figure \ref{fig:UNET1_Te1_DS1}, Figure \ref{fig:UNET2_Te1_DS2}, and Figure \ref{fig:UNET2_Te1_DS3} showcase random samples of input images, true and predicted masks by the UNet on test images of $Te_1$ and $Te_3$ of $DS_1$, $DS_2$, and $DS_3$. 

The very high performances of the network summarized in Table \ref{tab:segmentation_UNET1_performances} are reflected in the well-predicted masks in Figure \ref{fig:UNET1_Te1_DS1}. Such capabilities are also successfully transferred to the UNet developed afterward with multiple instances of boulders. As it is possible to see in Figure \ref{fig:UNET2_Te1_DS2}, the network is capable to predict correctly a large amount of boulders on the surface of Didymos with varying geometric and illumination conditions. It is also noted that the true masks exhibit challenging conditions in which a dense boulder's presence makes the surface almost entirely covered. This somehow reflects real environmental conditions, such as in Ryugu and Bennu. From the predicted masks in Figure \ref{fig:UNET2_Te1_DS2}, it is also possible to note an incorrect behavior of the network in those cases in which the body is not saturating the camera's FOV. In these cases, the network overestimates the presence of boulders over the edge, which ultimately drives down its performance. This seems to happen as the ray-tracing algorithm correctly labels true boulders over the projection of the edge in the image plane even when only a few pixels are observed over the very edge of the body, as it is also possible to observe from the true masks in Figure \ref{fig:UNET2_Te1_DS2}. Such a labeling mishap may have encouraged the network to actively label the entire edge as a boulder, which is a capability that is also transferred to the prediction on real images. Finally, it is also observed that the same phenomenon does not occur in the terminator region of the body, in which the boulder's labels are nullified by shadows.

Lastly, the final UNet is put to the test to predict boulders on real images from previously flown missions. It is remarked that these images have never been seen by the network during training and that they have only been seen in inference without further adjustments or fine-tuning on the network itself. First of all, as it is possible to see in Figure \ref{fig:UNET2_Te1_DS3}, it is commented that the noise levels from real cameras do not seem to pose a particular challenge to the network in terms of generalization. Albeit the network has been trained without a tailored noise setup to model any particular camera, the network performs similarly in all types of images considered. Is it thought that thanks to the injection of artificial noise as well as thanks to the particular care that has been put into the design of the $DS_1$ and $DS_2$ artificial datasets in their varying illumination, albedo, and scattering properties, the network is capable to generalize well enough. This is a promising feature for a direct network application on real sensor images. The only case which generated mild artifacts on images in $DS_3$ is visible in the 2nd case from the bottom in Figure \ref{fig:UNET2_Te1_DS3}.

As commented before, as it is possible to see from the true masks in Figure \ref{fig:UNET2_Te1_DS3}, too few boulders have been manually labeled to allow a quantitative assessment of the network in this case. Nonetheless, it is noted that boulder populations seem to be detected correctly over the surface. Comparing these types of predictions with the ones in \cite{Pugliatti2022_unet}, it is possible to conclude that the training presented in this work successfully pushed the architecture to detect well different-size boulders with a higher frequency than previously done. This ultimately proves to be a challenge on images of Bennu and Ryugu, since there is a risk of predicting boulder fields as uniformly spread features all over the surface. The desired behavior would be somewhat in between being able to predict small-medium size boulders on the surface as well as distinguish them clearly from large and prominent ones embedded in them. This ultimately poses a challenging problem that remains to be addressed for a real application.

\section*{Conclusions}
In this work, an IP pipeline to segment boulders on the surface of small body under a variety of illumination conditions has been developed through a 4-steps incremental training strategy. This task has been possible thanks to synthetically generated datasets of image-label pairs, which are made available in \cite{DatasetZenodoBoulders} . The IP network developed in this work exhibited excellent performance in segmenting isolated boulders. When applied to synthetic and real images of multiple boulders, the network has also demonstrated the capability to correctly isolate boulders, with degraded performance compared to single ones. The networks also exhibited high generalization capabilities, which are deemed to be delivered by the datasets intrinsic variabilities as well as by the addition of artificial noise on the images. 

Future iterations of this work will be directed toward an increase in the network generalization to a variety of small body shapes and boulder distributions. For this purpose, future updated versions of $DS_2$ may include different distributions for the same body as well as multiple global shape models. It has also been observed that labels of boulders on the edge of the body seem to introduce an erroneous behavior in the network. In future datasets iteration, it may be appropriate to remove such labels. The lack of labeled datasets for these types of IP tasks is ultimately a showstopper for the interested IP community for the development of data-driven algorithms. In this work, we have made all datasets but $DS_3$ publicly available for any interested user. It would be of interest to compare the performances of other approaches directly on the same datasets. A future collaborative effort could also be directed toward manual labeling of a sizeable chunk of real images obtained from previously flown missions. Finally, an open question remains on how to address the segmentation and identification of prominent boulders embedded into large boulder fields. This fractal nature represents a challenge in the labeling strategy presented in this work. 

\section*{Acknowledgment}
The authors would like to acknowledge the funding received from the European Union’s Horizon 2020 research and innovation programme under the Marie Skłodowska-Curie grant agreement No 813644. M.P. would also like to thanks Carmine Buonagura for the fruitful discussions and his help on the rendering of the datasets.

\vspace{3pt}
\bibliographystyle{ieeetr}
\titleformat{\section}[runin]{\normalsize\bfseries}{\thesection}{0em}{\addperiod}
{\footnotesize \bibliography{main}}

\newpage
\onecolumn

\end{document}